\documentclass[10pt,twocolumn,letterpaper]{article}

\usepackage[pagenumbers]{cvpr} 

\definecolor{cvprblue}{rgb}{0.21,0.49,0.74}
\usepackage[pagebackref,breaklinks,colorlinks,allcolors=cvprblue]{hyperref}
\usepackage{bm}
\usepackage{amsmath}
\usepackage{multirow}
\usepackage{indentfirst}
\usepackage{xcolor}
\usepackage{colortbl}
\def\paperID{*****} 
\def\confName{CVPR}
\def\confYear{2025}

\title{Dynamic Updates for Language Adaptation in Visual-Language Tracking}

\author{Xiaohai Li\textsuperscript{1}, Bineng Zhong\textsuperscript{1}\thanks{Corresponding Author}, Qihua Liang\textsuperscript{1}\thanks{Corresponding Author}, Zhiyi Mo\textsuperscript{2}, Jian Nong\textsuperscript{2}, Shuxiang Song\textsuperscript{1}\\
\textsuperscript{\rm 1}Key Laboratory of Education Blockchain and Intelligent Technology,\\ Ministry of Education,Guangxi Normal University, Guilin 541004, China\\
\textsuperscript{\rm 2}Guangxi Colleges and Universities Key Laboratory of Intelligent Software,\\ Wuzhou University, Wuzhou 543002, China\\
{\tt\small bruc\_0619@stu.gxnu.edu.cn, bnzhong@gxnu.edu.cn, qhliang@gxnu.edu.cn}\\ 
{\tt\small  zhiyim@gxuwz.edu.cn, nongjian@gxuwz.edu.cn,songshuxiang@mailbox.gxnu.edu.cn } \\
}

\begin{document}
\maketitle
\begin{abstract}
The consistency between the semantic information provided by the multi-modal reference and the tracked object is crucial for visual-language (VL) tracking. However, existing VL tracking frameworks rely on static multi-modal references to locate dynamic objects, which can lead to semantic discrepancies and reduce the robustness of the tracker. To address this issue, we propose a novel vision-language tracking framework, named DUTrack, which captures the latest state of the target by dynamically updating multi-modal references to maintain consistency.
Specifically, we introduce a Dynamic Language Update Module, which leverages a large language model to generate dynamic language descriptions for the object based on visual features and object category information. Then, we design a Dynamic Template Capture Module, which captures the regions in the image that highly match the dynamic language descriptions. Furthermore, to ensure the efficiency of description generation, we design an update strategy that assesses changes in target displacement, scale, and other factors to decide on updates. Finally, the dynamic template and language descriptions that record the latest state of the target are used to update the multi-modal references, providing more accurate reference information for subsequent inference and enhancing the robustness of the tracker.
DUTrack achieves new state-of-the-art performance on four mainstream vision-language and two vision-only tracking benchmarks, including LaSOT, LaSOT$_{\rm{ext}}$, TNL2K, OTB99-Lang, GOT-10K, and UAV123.
Code and models are available at https://github.com/GXNU-ZhongLab/DUTrack.

\end{abstract}

\begin{figure}[t]
  \centering
     \includegraphics[width=1\linewidth]{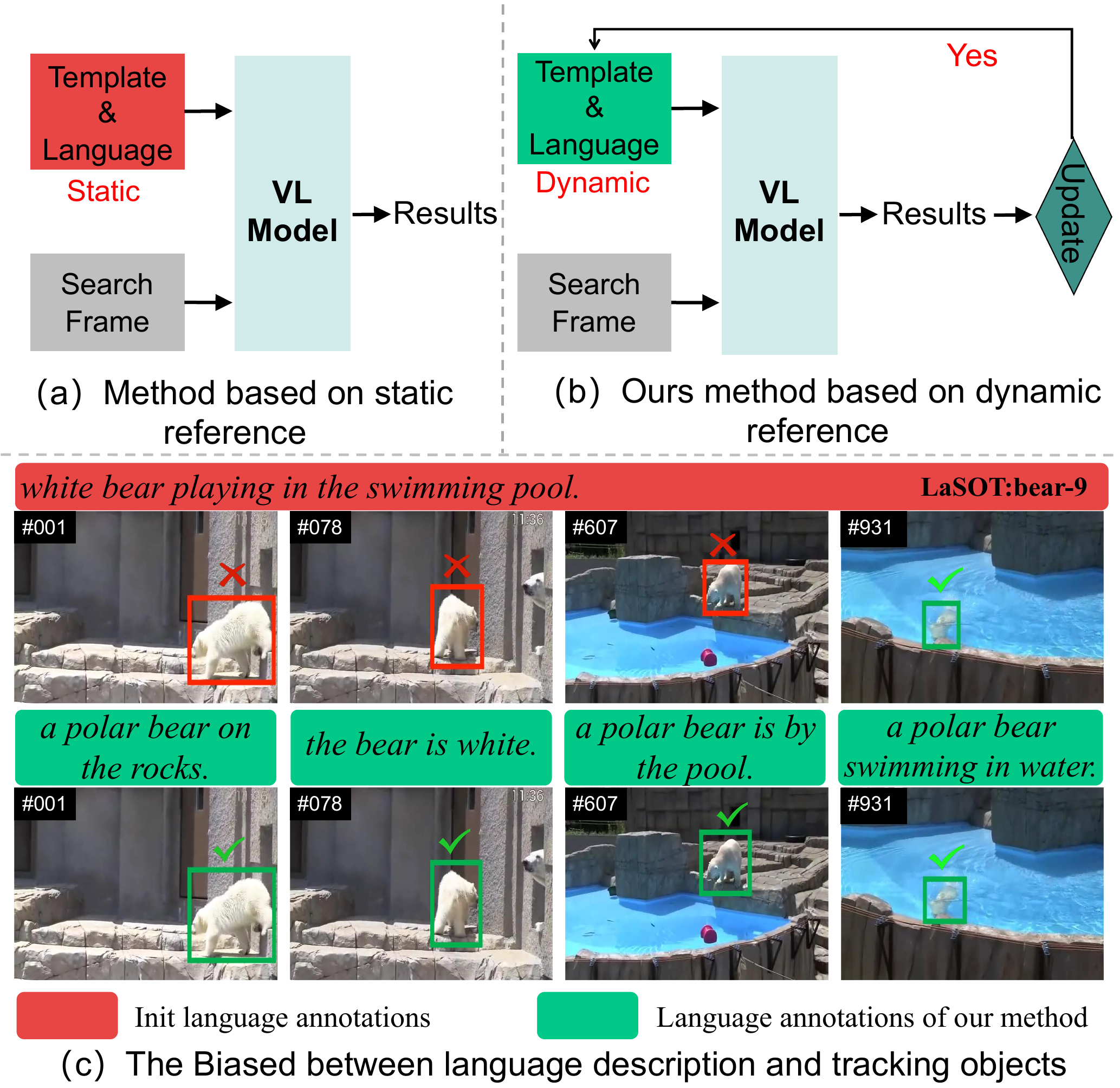}
    
       \caption{Comparison of different VL Tracking. (a) This vision-language tracking framework \cite{allinone,divertmore} relies on static multi-modal references. (b) Our proposes VL framework with dynamically updating multi-modal references. (c) Compare the semantic bias between static annotations and those generated by our method.
       }
       \label{figure:motivation}
       \vskip -0.1in
\end{figure}    
\section{Introduction}
\label{sec:intro}

Visual-language (VL) tracking aims to effectively track objects in video sequences based on multi-modal reference information provided by natural language descriptions and template frames. 
In real-world scenarios, objects may undergo appearance changes. 
These changes can cause discrepancies between the multi-modal reference information and the target, making the tracking task more complex. Therefore, maintaining the consistency between multi-modal target reference information and the target's state in the tracking scene during the tracking process has become a key challenge in this field.


Despite this challenge, a large number of current VL tracking methods \cite{feng2020,divertmore,UVLTrack,JointVL,mmtrack} tend to overlook it, instead relying on static multi-modal references, as shown in Fig.~\ref{figure:motivation}(a), and focusing on establishing stronger multi-modal interaction mechanisms. These methods can be categorized into one-stream and two-stream VL tracking frameworks. The two-stream framework \cite{divertmore,UVLTrack,JointVL,mmtrack} follows a typical three-stage paradigm: (\textit{i}) first, two independent encoders are used to extract uni-modal features from the initial language description and image, respectively; (\textit{ii}) then, multi-modal joint learning is employed to transfer the object information from the static multi-modal references to the search frame; (\textit{iii}) finally, a bounding box prediction head is used to output the tracking result. In contrast, the one-stream framework \cite{allinone,one_zhang,onevl,consistencies} differs in that it uses a unified encoder to simultaneously perform both multi-modal feature extraction and interaction, thereby simplifying the feature processing pipeline. 
These methods focus on designing effective multi-modal interaction mechanisms. Although they have achieved good performance, there is still a gap compared to the best vision-only trackers \cite{odtrack,aqatrack}. We believe the key reason for this gap lies in \textit{their excessive reliance on static multi-modal references composed of the initial template frame and language annotation}. As shown in Fig.~\ref{figure:motivation}(c), the initial language description can only provide the object's state at a specific moment and cannot continuously reflect the object's dynamic changes throughout the video sequence. 
Therefore, static multi-modal reference information easily diverges from the target's actual state, leading to lag or distortion in information during long-term tracking, which hampers accurate continuous localization and recognition of the object.

To address the above issues, we propose a novel VL tracking framework from the perspective of dynamically updating multi-modal reference information, as shown in Fig.~\ref{figure:motivation}(b). Our method can effectively reduce semantic discrepancies between natural language descriptions and the actual state of the target. Specifically, we design a Dynamic Template Capture Module (DTCM) and a Dynamic Language Update Module (DLUM) to update the visual and language references. The DTCM selects the top-k patches with the highest attention scores from the search image based on the attention map guided by the language annotations, and these patches represent the latest visual features of the object. The DLUM is based on a large language model. It generates updated language annotations using the search image and object category information. To improve update efficiency, we also design a strategy that adjusts the update frequency based on changes in the object's position, scale, and other factors. Extensive detailed experiments have demonstrated that our method, which combines dynamic visual and language information to update multi-modal references, can effectively enhance the performance of VL trackers. The major contributions of our work are summarized as follows:
\begin{itemize}
     \item We propose DUTrack, which enhances tracking capability by dynamically updating multi-modal references.
     \item We introduce two dynamic update modules, DTCM and DLUM, which can update visual and language reference.
     \item DUTrack sets a new state-of-the-art on four visual-language benchmarks and remains competitive on two vision-only benchmarks.
\end{itemize}

\begin{figure*}
    \centering
    \includegraphics[width=1.0\textwidth]{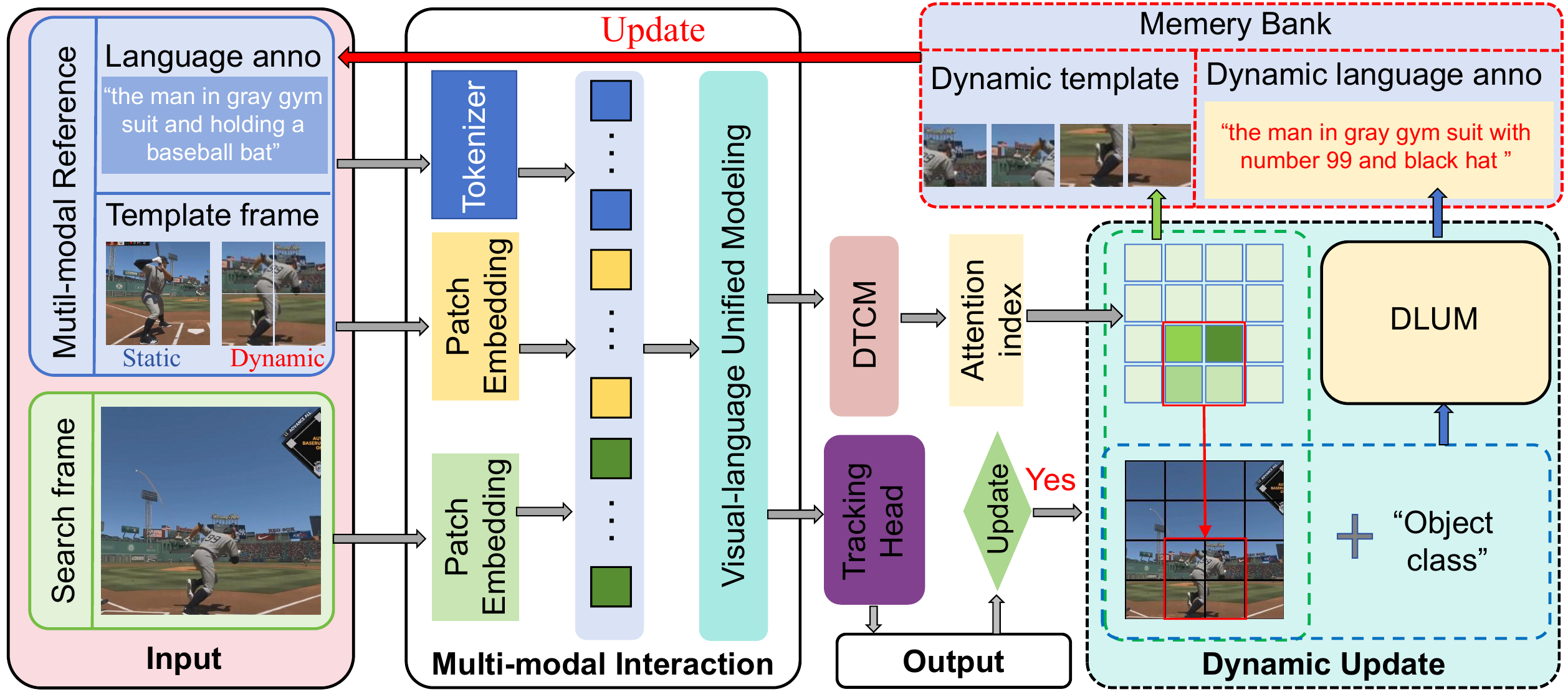}
    \caption{\textbf{Overall framework of the proposed DUTrack.} The input consists of two parts: search frame and multi-modal reference. The image and text information are transformed into tokens through Patch Embedding and Tokenizer processing, respectively. Then, these tokens enter the multi-modal interaction module for unified interaction. The resulting multi-modal features are processed through the tracking head to produce the final output. Based on this result, it is determined whether to update the multi-modal reference. The dynamic multi-modal reference is primarily responsible for generating a new reference according to the object’s state in the current frame.} 
    \label{fig:framework}
\end{figure*}

\section{Related Work}
\label{sec:Realted Work}
\textbf{Vision Tracking.} The vision trackers initialize their tracking procedure based on the given bounding box in the first frame. Based on whether the reference information is updated during tracking, trackers can be divided into two categories: static-reference trackers \cite{ aqatrack, HTtrack, ostrack, hu1,hu2} and dynamic-reference trackers \cite{evpt, odtrack, siamrdt, updataNet, stark}. Static-reference trackers rely heavily on the object's initial appearance features, focusing on modeling the relationship between the template frame and the search frame features. SiamFC \cite{simafc} extracts features from both the search frame and the template frame through a weight-sharing network and then uses a cross-correlation operation to propagate the reference information to the search frame features. OSTrack \cite{ostrack} integrates the feature extraction and feature fusion stages, significantly improving the performance of the tracker. The advantage of these static-reference trackers is their simplicity and relatively low computational cost, but their drawback is the difficulty in adapting to changes in the object's appearance. In contrast, dynamic-reference trackers continuously update the reference frame or object model over time to adapt to changes in the object’s appearance. \cite{updataNet} proposes an adaptive template update algorithm that updates the tracker’s template based on changes in the object's appearance. STARK \cite{stark} introduces dynamic reference updates and an update controller to capture changes in the object’s appearance. In this paper, we aim to propose a visual-language tracker that can automatically update multi-modal reference information, reducing the tracker’s dependence on the initial language annotations.

\textbf{Vision-language Tracking.} In contrast to traditional vision tracking tasks, vision language trackers not only utilize RGB reference information but also incorporate natural language descriptions as additional reference input. The primary objective is to integrate both visual and linguistic modalities to achieve more accurate object tracking. Current vision language trackers can be categorized into one-stream \cite{allinone,consistencies,one_zhang,onevl} and two-stream \cite{siamesevl,divertmore,JointVL,mmtrack} frameworks based on how they process multi-modal information.
Two-stream visual-language tracking frameworks typically use two different models to extract visual and language features (e.g., using ViT \cite{vit} to extract visual features and BERT \cite{bert} to extract language features), followed by multi-modal fusion to enable information interaction between the two modalities. UVLTrack \cite{UVLTrack} extracts features from both modalities separately at a shallow level and then performs feature fusion at a deeper level. The VLT \cite{divertmore} introduces a modality mixer during the independent extraction of different modality features, effectively promoting multi-modal interaction. In contrast, one-stream trackers use only one encoder to simultaneously extract features from both modalities and perform multi-modal feature fusion in a single stage. To enhance the object perception of visual-language trackers in complex scenes, Zhang \cite{one_zhang} proposed an integrated framework that employs a unified transformer backbone to jointly learn feature extraction and interaction. ATTracker \cite{consistencies} uses an asymmetric multi-encoder to unify the learning of multi-modal features. 

Although the aforementioned visual-language trackers have achieved significant success, they still have shortcomings: they rely on static multi-modal references. These multi-modal references consist of an initial template frame and language annotation. The initial language annotation typically provides an overview of the object's complete actions or describes its state at a specific moment. Similarly, the initial template frame describes the target's visual state at the beginning. Relying solely on static references can lead to discrepancies between the references and the target's state. In contrast, we propose a visual-language tracking framework that can dynamically update multi-modal references to maintain consistency between the references and the object.

\section{Method}
In this section, we will provide a detailed introduction to the proposed DUTrack. First, we review the tracking process of DUTrack. Then, we introduce the multi-modal interaction module in detail. Finally, we delve into the dynamic template capture module and the dynamic language update module.

\subsection{Overview}
An overview of the DUTrack framework is shown in Fig.~\ref{fig:framework}. This framework is very concise and consists of four main components: the multi-modal interaction module, the dynamic update module, the dynamic change capture module, and the tracking head.
The input to this pipeline primarily consists of two parts: the search image \({S}\in\mathbb{R}^{3\times H_S\times W_S}\) and the multi-modal reference, where the multi-modal reference is composed of the template frame \({T}\in\mathbb{R}^{3\times H_T\times W_T}\) and language annotations \({L}\). \(H\) and \(W\) represent the height and width of the image. After the input enters the multi-modal interaction module, \(S\) and \(T\) are transformed into tokens through image patch embedding operations. \(L\) is converted into a string of tokens through a tokenizer. Then, we concatenate these tokens and input them into the visual-language unified modeling for multi-modal interaction. In the final stage of interaction, the module outputs the search feature map and the global attention map. 
The global attention map, processed through the dynamic template capture module, can index the tokens in \({S}\) that have a high match with \({L}\). 
The search feature map is input into the tracking head to obtain results. 
By analyzing changes in the object's position, scale, and other attributes, the search image and object category information are fed into the dynamic language update module to generate dynamic language annotations. Finally, the dynamic template and language annotations update the multi-modal reference.


\subsection{Multi-modal Interaction}
Existing visual tracking frameworks typically use two separate encoders to independently extract features from different modalities before fusing them. This approach results in a lack of deep correlations between the extracted features, leading to sub-optimal performance in complex scenarios. Inspired by recent advancements in joint feature learning and relation modeling, we adopt a one-stream framework for multi-modal feature learning. To capture rich spatial information, we use HiViT \cite{hivit} for unified visual-language modeling. Unlike the vanilla ViT \cite{vit}, which directly uses \(16\times16\) embeddings, the search image and template image are transformed into tokens \({S}_t\in\mathbb{R}^{N_S\times D}\) and \({T}_t\in\mathbb{R}^{N_T\times D}\) through three stages of down-sampling ( a \(4\times4\) embedding layer and two \(2\times2\) merging layers). The language annotation \(L\) is converted into \({L}_t\in\mathbb{R}^{N_L\times D}\) through BERT's \cite{bert} tokenizer, and \({L}_t\) begins with a \([CLS]\) token. Here \(N\) represents the number of tokens, and  \(D\) represents the dimensionality, \(N_{S}=H_{S}W_{S}/16^{2}\), \(N_{T}=H_{T}W_{T}/16^{2}\), \(N_{L}=16\), \(D=512\). It is worth noting that \({T}_t\) includes not only the initial template patches but also the patches from the dynamic template. Formally, the operation of unified visual-language modeling can be expressed as:
\begin{equation}\begin{aligned}
\label{eq:qkv}
{Q}={K}&={V}=[{L}_t;{T}_t;{S}_t], \\
feat,attn&={MHSA}(Q,K,V), \\
f_{vl}=[{L}_t;&{T}_t;{S}_t]+{LN}(\lambda_{1}\cdot feat), \\
[{F}_L;{F}_T;{F}_S]&=f_{vl}+{LN}(\lambda_{2}\cdot MLP(f_{vl})),
\end{aligned}\end{equation}
where \(LN\) represents the layer normalization, \(MHSA\) stands for multi-head self attention and \([;]\) denotes the concatenation operation. \(\lambda_{1}\) and \(\lambda_{2}\) are two learnable parameters. Finally, the search feature \({F}_S\) is fed into the tracking head to obtain the result \([x,y,w,h]\).

\subsection{Dynamic Template Capture Module}
\begin{figure}[t]
      \centering
       \includegraphics[width=1\linewidth]{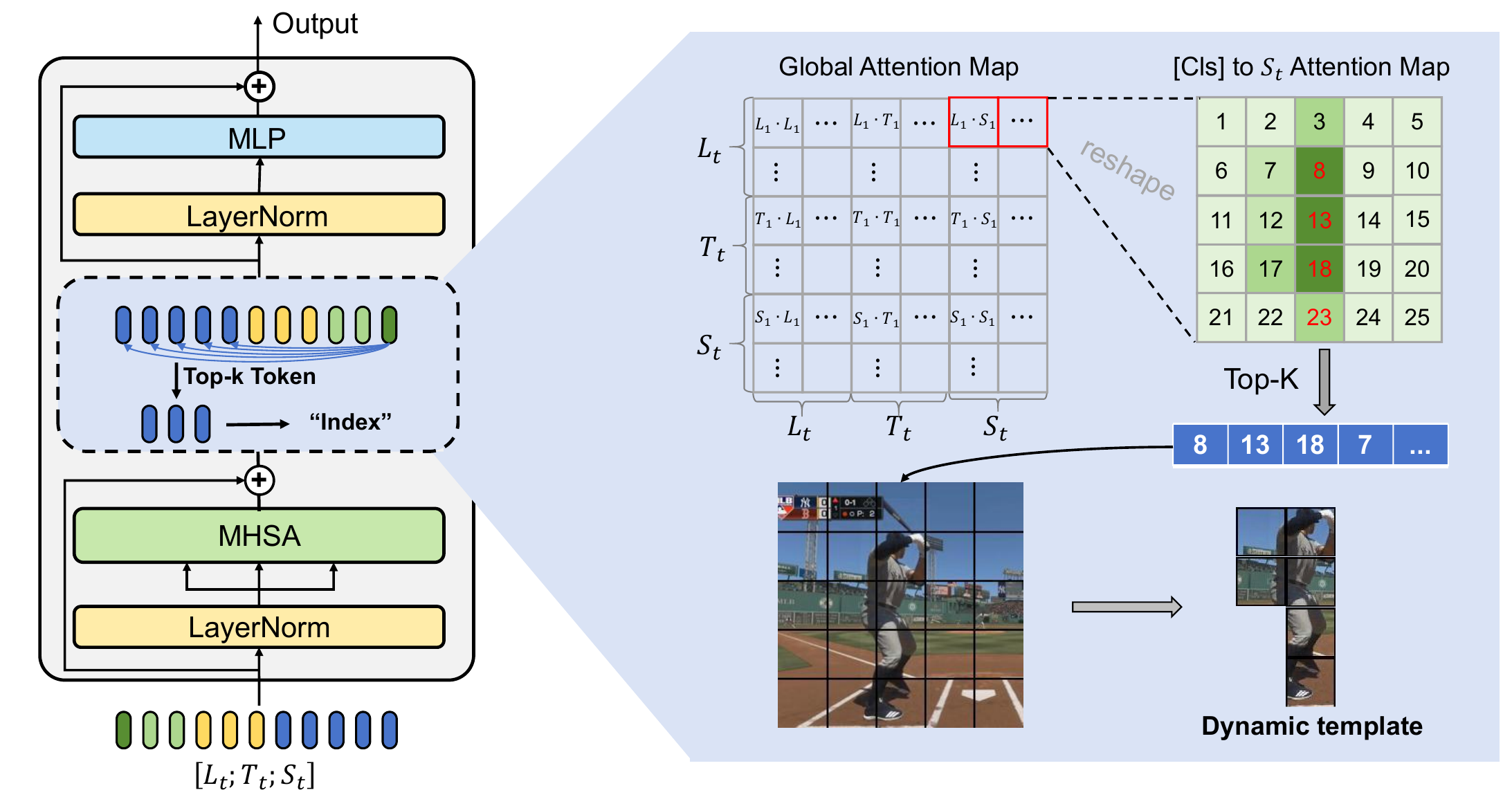}
    
       \caption{Illustration of the process of capturing dynamic templates. The left side shows the unified visual-language modeling generating a global attention map, while the right side captures dynamic templates based on the global attention map.}
       \label{figure:dcm}
\end{figure}

The object appearance information provided by a static template frame only captures the object's appearance at a specific moment, which is insufficient to offer the tracker adequate spatial cues. To address this issue, we designed a simple yet efficient dynamic template capture module. This module extracts high-response patches from the search image and uses them as dynamic templates for the subsequent frame. Specifically, the multi-head self-attention operation can be seen as spatial aggregation of tokens with normalized importance, as shown in Fig.~\ref{figure:dcm}, This is measured by the dot product similarity between each pair of tokens. The calculation for each token is as follows:
\begin{equation}\begin{aligned}
\label{eq:attn}
&{A=Softmax}(\frac{{QK}^{T}}{\sqrt{{d}}})\cdot{V}, \\
\end{aligned}\end{equation}
Where \(A\) is the similarity matrix between tokens, based on Eq.~\ref{eq:qkv}, Eq.~\ref{eq:attn} can be extended as:
\begin{equation}\begin{aligned}
\label{eq:attn2}
&A={Softmax}(\frac{[{Q}_{L};{Q}_{T};{Q}_{S}][{K}_{L};{K}_{T};{K}_{S}]^{T}}{\sqrt{{d}}})\cdot[{V}_{L};{V}_{T};{V}_{S}],
\end{aligned}\end{equation}
Where subscripts \(L\), \(T\), and \(S\) denote matrix items belonging to language annotations, templates, and search regions. The module we designed aims to identify patches in the search area that highly match the language annotations. The \([CLS]\) token in the language annotations can comprehensively summarize the semantic information. Therefore, we select the attention map of \([CLS]\) towards the search area from \(A\), which can be represented as follows:
\begin{equation}\begin{aligned}
\label{eq:attn2}
&A_{l2s}={Softmax}(\frac{[{Q}_{CLS}][{K}_{S}]^{T}}{\sqrt{{d}}})\cdot[{V}_{L};{V}_{T};{V}_{S}].
\end{aligned}\end{equation}
Then, we select the top-k patches with the highest similarity from \(A_{l2s}\) and record their indices. Based on these indices, we locate the corresponding regions in the image to serve as dynamic templates.

\subsection{Dynamic Language Update Module}
The official annotations provided by vision-language tracking benchmarks are limited to describing the short-term state of the object, which poses challenges to frame-by-frame reasoning in tracking paradigms. To enhance the contribution of language annotations to tracking performance, we leverage large language models to dynamically generate descriptions of the target at specific time points. Typically, these annotations include semantic information about the object's position, scale, and color. Based on this, we designed a dynamic update strategy. 
Specifically, we introduce an object stamp \(r_{stamp}:[x_1,y_1,w_1,h_1]\) that records target information from the last update, initialized with annotations from the target's first frame. The tracking result for the current frame is also represented as  \(r_i:[x_2,y_2,w_2,h_2]\). By comparing the displacement of the center point, changes in the target's scale, and alterations in the mean color within the bounding box between \(r_{stamp}\) and \(r_i\), we dynamically determine whether the language annotations should be updated. This process can be described as follows:
\begin{equation}
    \label{eq:s}
    \Delta S=\frac{w_\mathrm{1}\times h_\mathrm{1}}{w_\mathrm{2}\times h_\mathrm{2}},
\end{equation}
\begin{equation}
   \label{eq:d}
    \Delta D=\sqrt{(x_\mathrm{1}-x_\mathrm{2})^2+(y_\mathrm{1}-y_\mathrm{2})^2},
\end{equation}
\begin{equation}
    \Delta C=\sqrt{(R_{\mathrm{1}}-R_{\mathrm{2}})^2+(G_{\mathrm{1}}-G_{\mathrm{2}})^2+(B_{\mathrm{1}}-B_{\mathrm{2}})^2},
\end{equation}
where \(\Delta S\), \(\Delta D\), and \(\Delta C\) represent the changes in scale, position, and color mean of the current target relative to the object stamp. \(R\), \(G\), and \(B\) represent the mean values of the RGB pixels, respectively. Then, we manually set three thresholds to control the update frequency. Finally, we leverage a large language model to simultaneously learn the relationship between the search image and the object category, effectively generating contextually relevant descriptions associated with the object category.
    

\section{Experiments}
\label{sec:experiments}
\subsection{Implementation Details}
\textbf{Model.} We use BERT \cite{bert}'s tokenizer and the hierarchical patch embedding method from HiViT \cite{hivit} to convert both language and images into tokens. Then, we initialize our ViT module as a unified vision-language encoder using Fast-ITPN \cite{fast-itpn}. The one-stream framework can effectively improve the efficiency of multi-modal interaction. As shown in Tab.~\ref{tab:speed}, under the premise of using the same ViT-base as the backbone, our tracker demonstrates significant competitiveness in both speed and AUC performance. 

\begin{table}[h]
   
    \centering
    \caption{Comparison of performance, model parameters, and inference speed on TNL2K \cite{tnl2k}.}
    \label{tab:speed}
    \fontsize{9}{10.8}\selectfont
    \resizebox{0.48\textwidth}{!}
    {
    \begin{tabular}{c|cccccc}
     \toprule
     \multicolumn{1}{c|}{Tracker} & \multicolumn{1}{c}{AUC(\%)$\uparrow$} 
     & \multicolumn{1}{c}{Params(M)$\downarrow$} & \multicolumn{1}{c}{Speed(fps)$\uparrow$}& \multicolumn{1}{c}{Device} \\
     \midrule
     JiontNLT \cite{JointVL} & 56.9 & 153.0 & 25.6 & RTX-2080Ti \\
     MMTrack \cite{mmtrack} & 58.6 & 176.9 & 36.2 & RTX-2080Ti \\
     Ours & \textbf{64.9} & \textbf{69.9} & \textbf{43.5} & RTX-2080Ti \\

     \bottomrule

    \end{tabular}
    }
    
    \label{tab:model}
\end{table}
\textbf{Training and Inference.}
Based on different search frame input sizes, we trained two models, DUTrack-256 and DUTrack-384. The training process is as follows:
we first use LaSOT \cite{LaSOT}, GOT-10K \cite{got10k}, COCO \cite{coco}, TrackingNet \cite{trackingnet}, and TNL2k \cite{tnl2k} for the first stage training. In this stage, we do not use language information, aiming to develop strong visual tracking capability. We employ AdamW to optimize the network parameters, with both the learning rate and weight decay set to $1 \times 10^{-4}$. During this stage, we train for 150 epochs, with a sample size of $60,000$ images. In the second stage, we use LaSOT \cite{LaSOT}, GOT-10K \cite{got10k}, and TNL2K \cite{tnl2k} as the training benchmarks and introduce a dynamic update multi-modal reference mechanism. Since the training strategy is based on random sampling, to reduce training time, we directly use the language annotations provided by DTLLM-VLT \cite{dtllm} as our input. This stage involves 50 epochs of training, with the same training parameters as in the first stage. During inference, the top-k is set to 3, and the large language model used is BLIP \cite{blip}.

\begin{table*}
    \centering
    \caption{Performance comparison on four benchmarks, including LaSOT \cite{LaSOT}, LaSOT$_{\rm{ext}}$ \cite{LaSOText}, TNL2K \cite{tnl2k}, and OTB99-Lang \cite{otblang}. We compare DUTrack with state-of-the-arts. These works can be mainly divided into visual-only trackers and VL trackers. The top three results are highlighted in {\color{red}red} and {\color{blue}blue}, respectively. }
    \label{tab:results}
    \resizebox{\textwidth}{!}{
  \begin{tabular}{c|l|c|ccc|ccc|ccc|cc}
    \toprule
     \multicolumn{1}{c|}{\multirow{2}{*}{Type}} & \multicolumn{1}{c|}{\multirow{2}{*}{Method}} & \multicolumn{1}{c|}{\multirow{2}{*}{Source}}
      & \multicolumn{3}{c|}{LaSOT} &\multicolumn{3}{c|}{LaSOT$_{\rm{ext}}$} & \multicolumn{3}{c|}{TNL2K} & \multicolumn{2}{c}{OTB99-Lang} \\ \cline{4-14}
      & & & AUC & P${_{\rm{Norm}}}$ & P & AUC & P${_{\rm{Norm}}}$ & P & AUC & P${_{\rm{Norm}}}$ & P & AUC & P \\
      \midrule
      \multicolumn{1}{c|}{\multirow{15}{*}{\rotatebox{90}{Vision-only}}} 
      & SiamFC \cite{simafc} & ECCV2016 & 33.6 & 42.0 & 33.9 & 23.0 & 31.1 & 26.9 & 29.5 & 45.0 & 28.6  & - & - \\
      & SiamBAN \cite{siamBAN} & CVPR2020 & 51.4 & 59.8 & 52.1 & - & - & - & 41.0 & 48.5 & 41.7  & - & - \\
      & TransT \cite{chen2021transformer} & CVPR2021 & 64.9 & 73.8 & 69.0 & - & - & - & 50.7 & 57.1 & 51.7  & - & - \\
      & Stark \cite{stark} & ICCV2021 & 67.1 & 77.0 & - & - & - & -  & - & - & - & - & - \\
      & GTELT \cite{GTELT} & CVPR2022 & 67.7 & - & - & 45.0 & 54.2 & 52.2  & - & - & - & - & - \\
      & TransInMo \cite{TransInMo} & IJCAI2022 & 65.7 & 76.0 & 70.7 & - & - & - & 52.0 & 58.5 & 52.7  & - & - \\
      & OSTrack-256 \cite{ostrack} & ECCV2022 & 69.1 & 78.7 & 75.2 & 47.4 & 57.3 & 53.3  & 54.3 & - & - & - & - \\
      & OSTrack-384 \cite{ostrack} & ECCV2022 & 71.1 & 81.1 & 77.6 & 50.5 & 61.3 & 57.6  & 55.9 & - & - & - & - \\
      & SeqTrack-B256 \cite{seqtrack} & CVPR2023 & 69.9 & 79.7 & 76.3 & 49.5 & 60.8 & 56.3 & 54.9 & - & - & - & - \\
      & SeqTrack-B384 \cite{seqtrack} & CVPR2023 & 71.5 & 81.1 & 77.8 & 50.5 & 61.6 & 57.5 & 56.4 & - & - & - & - \\
      & AQATrack-256 \cite{aqatrack} & CVPR2024 & 71.4 & 81.9 & 78.6 & 51.2 & 62.2 & 58.9 & 57.8 & - & 59.4 & - & - \\
      & AQATrack-384 \cite{aqatrack} & CVPR2024 & 72.7 & 82.9 & 80.2 & 52.7 & 64.2 & 60.8 & 59.3 & - & 62.3 & - & - \\
      & ODTrack-B384 \cite{odtrack} & AAAI2024 & 73.2 & 83.2 & 80.6 & 52.4 & 63.9 & 60.1 & 60.9 & - & - & - & - \\
      \hline
      \multicolumn{1}{c|}{\multirow{10}{*}{\rotatebox{90}{Vision-Language}}} 
      & LSTMTrack \cite{lstmtrack} & WACV2020 & 35.0 & - & 35.0 & - & - & - & 25.0 & 33.0 & 27.0 & 61.0 & 79.0 \\ 
      & SNLT \cite{feng2021siamese} & CVPR2021 & 54.0 & 63.6 & 57.4 & - & - & - & - & - & - & 66.6 & 84.8 \\ 
      & GTI \cite{GTI} & TCSVT2021 & 47.8 & - & 47.6 & - & - & - & - & - & - & 58.1 & 73.2 \\ 
      & TNL2K-II \cite{tnl2k} & CVPR2021 & 51.3 & - & 55.4 & - & - & - & 42.0 & 50.0 & 42.0 & 68.0 & 88.0 \\ 
      & TransVLT \cite{transvl} & PRL2023 & 66.4 & - & 70.8 & - & - & - & 56.0 & 61.7 & -  & 69.9 & {91.2} \\ 
      & JointNLT \cite{JointVL} & CVPR2023 & 60.4 & 69.4 & 63.6 & - & - & - & {56.9} & {73.6} & {58.1} & 65.3 & 85.6 \\
      & MMTrack-384 \cite{mmtrack} & TCSVT2023 & {70.0} & {82.3} & {75.7} & {49.7} & {59.9} &{55.3} & {58.6} & {75.2} & {59.4} & {70.5} & {91.8} \\
      & ATTrack \cite{ATTrak} & MM2024 & {63.7} & {-} & {67.3} & {-} & {-} &{-} & {56.9} & {75.0} & {64.7} & {69.3} & {90.3} \\
      & OSDT \cite{one_zhang} & TCSVT2024 & {64.3} & {68.6} & {73.4} & {-} & {-} &{-} & {59.3} & {-} & {61.5} & {66.2} & {86.7} \\
      & UVLTrack-B \cite{UVLTrack} & AAAI2024 & {69.4} & {-} & {74.9} & {49.2}  &{-} & {55.8} & {62.7} & {-} & {65.4} & {60.1} & {79.1} \\
      & UVLTrack-L \cite{UVLTrack} & AAAI2024 & {71.3} & {-} & {78.3} & \color{blue}{51.2}  &{-} & {57.6} & {64.8} & {-} & {68.8} & {63.5} & {83.2} \\
      & QueryNLT \cite{queryNLT} & CVPR2024 & {59.9} & {69.6} & {63.5} & {-} & {-} &{-} & {57.8} & {75.6} & {58.7} & {66.7} & {88.2} \\

      \cline{2-14}
      & DUTrack-256 & Ours & \color{blue}{73.0} & \color{blue}{83.8} &\color{blue}{81.1} & {50.5} & \color{blue}{61.5} & \color{blue}{58.1} & \color{blue}{64.9} & \color{blue}{82.9} & \color{blue}{70.6}  & \color{blue}{70.9} & \color{blue}{93.9} \\
      & DUTrack-384 & Ours & \color{red}{74.1} & \color{red}{84.9} &\color{red}{82.9} & \color{red}{52.5} & \color{red}{63.6} & \color{red}{60.5} & \color{red}{65.6} & \color{red}{83.2} & \color{red}{71.9} & \color{red}{71.3} & \color{red}{95.7} \\
    \bottomrule
    \end{tabular} }
\end{table*}

\subsection{State-of-the Art Comparisons}
In this section, we evaluated DUTrack's performance on four vision-language tracking benchmarks, including TNL2K \cite{tnl2k}, LaSOT \cite{LaSOT}, LaSOT$_{\rm{ext}}$ \cite{LaSOText}, and OTB99-Lang \cite{otblang}. The results are shown in Tab.~\ref{tab:results}. Furthermore, since DUTrack has the ability to generate language descriptions, it can also be evaluated on vision-only benchmarks that do not provide language annotations. We additionally evaluated it on the vision-only benchmarks GOT-10K \cite{got10k} and the drone tracking dataset UAV123 \cite{uav123}. Their results are shown in Tab.~\ref{tab:got10k&uav}.

\textbf{LaSOT} is a large-scale vision-language tracking benchmark, where the language annotations primarily describe the object's behavior and state throughout the video. It consists of 1,400 video sequences, with the training and testing sets split in a 1,220/280 ratio. As shown in Tab.~\ref{tab:results}, our proposed DUTrack-256 outperforms the QueryNLT \cite{queryNLT}, published at CVPR 2024, with improvements of 14\%, 14.2\%, and 17.6\% in AUC, \(P_{Norm}\), and P, respectively. DUTrack-384 achieves new state-of-the-art results with an AUC  of 74.1\%, a \(P_{Norm}\) of 84.9\%, and a P  of 82.9\%. Compared to the top vision-only tracker ODTrack-B384, we achieve improvements of 0.9\%, 1.7\%, and 2.3\% on these three metrics. LaSOT is a benchmark with relatively long average sequences, where the mismatch between the language annotations and the frames is more pronounced, which has prevented previous VL trackers from surpassing top vision-only trackers on this benchmark. However, DUTrack’s performance demonstrates that our proposed dynamic update mechanism for multi-modal reference information significantly advances vision-language tasks.

\textbf{LaSOT$_{\rm{ext}}$} is an extension of the LaSOT dataset, containing 150 sequences. As shown in Tab.~\ref{tab:results}, we observe that our DUtrack-256 achieves comparable results of 50.5\%, 61.5\%, and 58.1\% in terms of success, P$_{\rm{Norm}}$, and precision score. Compared to UVLTrack \cite{UVLTrack} with the same input size, we outperform by 1.3\%, 1.6\%, and 2.8\% on the three metrics, respectively. DUTrack-384 outperforms MMTrack-384 \cite{mmtrack} with an improvement of 2.8\% in AUC, 3.7\% in P$_{\rm{Norm}}$, and 5.2\% in P. Compared to the top vision-only trackers, DUTrack's performance is on par. We believe the lack of significant improvement is due to the relatively small test dataset in this benchmark, which leads to considerable fluctuations in the experimental results.

\begin{table*}[h]
    \centering
    \caption{Comparison with state-of-the-art methods on GOT-10k and UAV123 benchmarks in AO and AUC score.We add a symbol * over GOT-10k to indicate that the corresponding models are only trained with the GOT-10k training set.}
    \resizebox{1\textwidth}{!}{
    \begin{tabular}{c|cccccccccc|cc}
     \toprule
     \multicolumn{1}{c|}{} & \multicolumn{1}{c}{SiamFC}& \multicolumn{1}{c}{MDNet} & \multicolumn{1}{c}{Ocean} & \multicolumn{1}{c}{SiamPRN++}
     & \multicolumn{1}{c}{TrDimp} & \multicolumn{1}{c}{TransT}&  SimaTrack &VideoTrack &SeqTrack&ODTrack& \multicolumn{1}{|c}{DUTrack-256} &\multicolumn{1}{c}{DUTrack-384} \\
     \midrule
    GOT-10k$^*$& 34.8&29.9& 61.1&51.7 &67.1 &67.1 &68.6 &72.9 &74.7 &\color{blue}77.0 & 76.7& \color{red}77.8\\
     UAV123& 46.8&52.8& 57.4&61.0 &69.1 &69.1 &\color{blue}69.8 &69.7 &69.2 &- & 69.3& \color{red}70.1\\
    \bottomrule 
     
    \end{tabular}
    }
    
    \label{tab:got10k&uav}
\end{table*}

\textbf{TNL2K} is a tracking benchmark specifically designed for vision-language tracking. Similar to LaSOT, it only provides initial language annotations. As shown in Tab.~\ref{tab:results}, Our approach gets the best scores of 65.6\%, 83.2\%, and 71.9\% in terms of AUC, P$_{\rm{Norm}}$, and precision, respectively. Moreover, our DUTrack-256 uses a base encoder, yet it still slightly outperforms UVLTrack-L \cite{UVLTrack}, which uses a large encoder with the same input size, across all three metrics. Compared to ODTrack \cite{odtrack}, which currently holds the best scores on this benchmark, our DUTrack-384 achieves a 4.6\% improvement in AUC. Although the TNL2K dataset provides high-quality language annotations, these annotations are fixed and cannot offer more closely matched target information. This limitation hinders previous methods from achieving optimal performance. At the same time, it clearly demonstrates that our dynamic language annotation updates can significantly enhance the performance of current vision-language trackers.

\textbf{OTB99-Lang} is an extension of the OTB benchmark in terms of language description. Tracker performance is evaluated using the area under the curve (AUC) and precision (P) metrics, following the protocol established by the official OTB evaluation. As shown in Tab.~\ref{tab:results}, DUTrack-256 produced results that were competitive, with success and precision scores of 70.9\% and 93.9\%, respectively.  DUTrack-384 gets the best scores of 71.3\%, 95.7\%, and 71.9\% in terms of AUC and precision, respectively.

\begin{figure}
    \centering
    \includegraphics[width=0.45\textwidth]{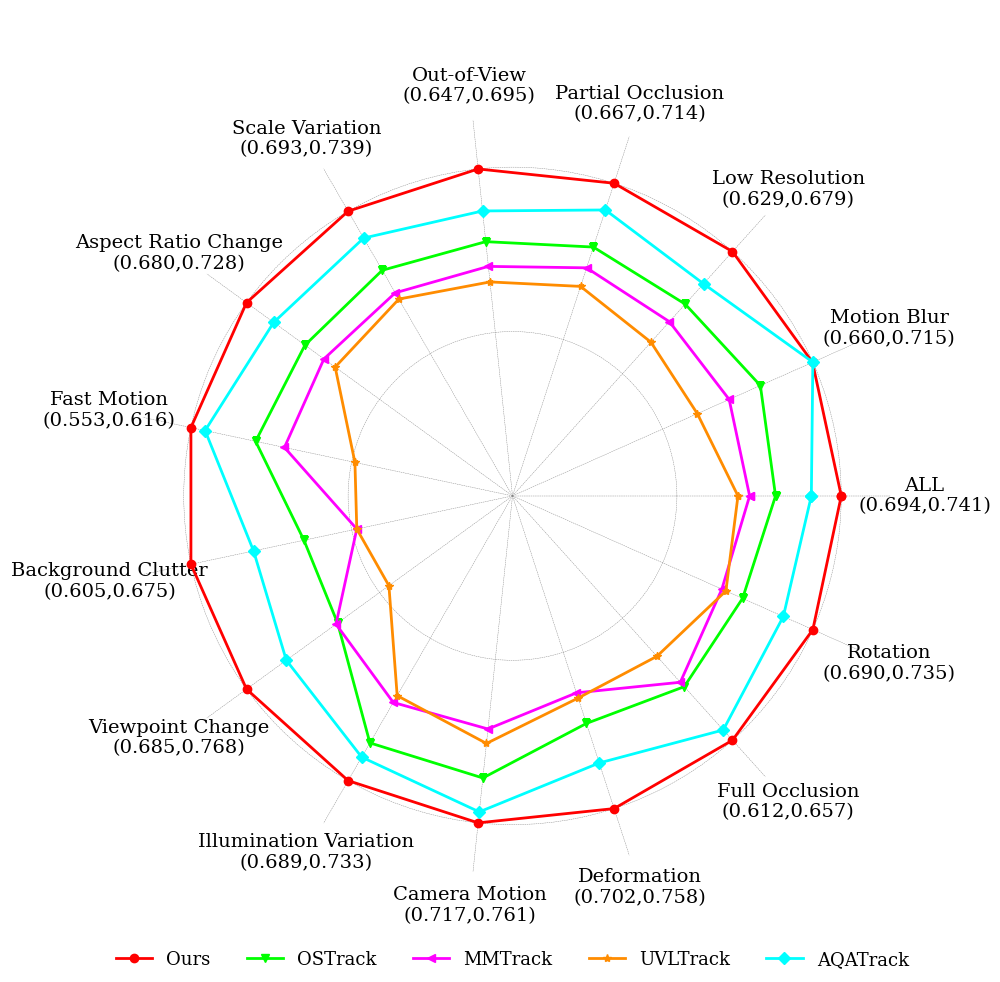}
    \caption{Attribute-based evaluation on the LaSOT test set. AUC score is used to rank different trackers.}
    \label{fig:eao_lasot}
\end{figure}

\textbf{GOT-10k} is a vision-only tracking benchmark, and its test set does not provide any language annotations. The dataset is evaluated using two metrics: Average Overlap (AO) and Success Rate (SR). As shown in Tab.~\ref{tab:got10k&uav}, DUTrack-256 achieved AO 76.7\%, which shows an improvement of 2.0\% compared to SeqTrack \cite{seqtrack}. Compared to ODTrack \cite{odtrack}, DUTrack-384 exceeds it by 0.8\% in AO.

\textbf{UAV123} is a benchmark for object tracking research, focusing on video sequences captured by drones, with its main evaluation metric being AUC. From Tab.~\ref{tab:got10k&uav}, we can see that the two versions of DUTrack achieve AUC scores of 69.3\% and 70.1\%, respectively. The improvement over SimTracker is relatively small, only 0.2\%. We believe that language descriptions provide limited assistance for small targets captured by low-altitude drones.

\begin{figure}
    \centering
    \includegraphics[width=0.48\textwidth]{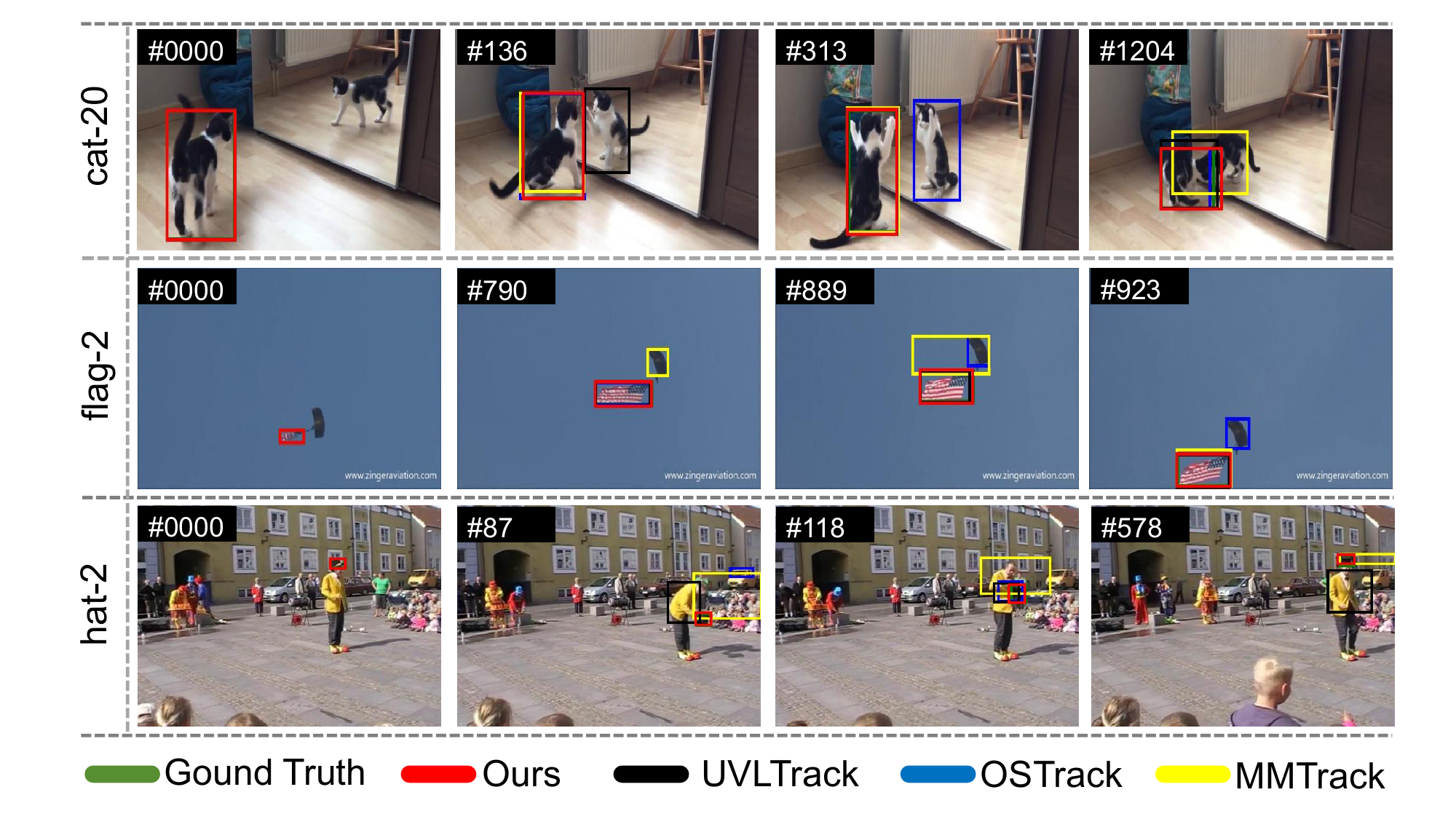}
    \caption{Qualitative comparison results of our tracker with two VL trackers(i.e UVLTrack and MMtrack) and one visual-only tracker OSTrak on three challenging sequences from the LaSOT benchmark. Better viewed in color with zoom-in.} 
    \label{fig:vis_box}
\end{figure}

\subsection{Ablation Study}
In this section, we provide detailed experiments to verify the impact of our proposed dynamic update of multi-modal reference information on performance. We use DUTrack-256 without any multi-modal reference information update as the baseline for this experiment. Then, we incrementally add modules to this baseline and conduct more granular comparative experiments on each module.
\begin{table}[h]
   
    \centering
    \caption{Study on different top-k for DTCM.}
    \label{tab:study_DC}
    \fontsize{8}{7}\selectfont
    \resizebox{0.48\textwidth}{!}
    {
    \begin{tabular}{c|c|ccc|cc}
     \toprule
     \multicolumn{1}{c|}{\multirow{2}{*}{Num}} &\multicolumn{1}{c|}{\multirow{2}{*}{top-k}} & \multicolumn{3}{c|}{LaSOT} 
     & \multicolumn{2}{c}{GOT-10K}  \\ \cline{3-7}
     & &AUC & P$_{\rm{Norm}}$ & P &AO & SR$_{\rm{0.5}}$\\
     \midrule
     \#1&0 & 71.0 & 79.7 & 75.9 & 72.2 &82.8 \\
     \#2&1 & 71.1 & 79.2 & 77.6 & 73.4 &83.9 \\
     \#3&2 & 71.5 & 81.0 & 77.8 & 74.2 &85.0 \\
     \#4&3 & 71.7 & 81.9 & 78.1 & 74.6 &84.7 \\
     \bottomrule
    \end{tabular}
    }
    \vskip -0.1in
    \label{tab:model}
\end{table}

\textbf{Study on the  Dynamic Template Capture Module.} To analyze the impact of the DTCM on experimental results, we use the number of top-k patches where the semantic features of the image and language annotations match most closely as a variable. We evaluate the performance of different top-k values on vision-language and vision-only benchmarks. As shown in Tab.~\ref{tab:study_DC}, when top-k is set to 0, it represents the baseline performance. As the top-k number increases, the AUC on LaSOT increases from 71.0\% to 71.7\%, and the AO on GOT-10k improves from 72.2\% to 74.6\%. The notable AUC improvements on both vision-language and vision-only benchmarks clearly indicate that DTCM enhances the tracker's perception of target appearance by capturing new object spatial features.
   
\begin{table}[h]
    \centering
    \caption{Study on different update parameters for DLUM}
    \label{tab:study_DL}
    \fontsize{8}{6}\selectfont
    \resizebox{0.48\textwidth}{!}{
    \begin{tabular}{c|c|c|ccc}
     \toprule
     \multicolumn{1}{c|}{\multirow{2}{*}{Num}} & \multicolumn{1}{c|}{\multirow{2}{*}{$\Delta S$}} & \multicolumn{1}{c|}{\multirow{2}{*}{$\Delta D$}} & \multicolumn{3}{c}{LaSOT} \\ 
     \cline{4-6}
     & & & AUC & P$_{\rm{Norm}}$ & P \\
     \midrule
     \#1 & 0 & $16\times \mathrm{stride}$ & 72.4 & 83.2 & 80.3  \\
     \#2 & 0.5 & $2\times \mathrm{stride}$ & 72.5 & 83.2 & 80.4  \\
     \#3 & 0.8 & $1\times \mathrm{stride}$ & 72.7 & 83.4 & 80.6  \\
     \#4 & 1 & $0\times \mathrm{stride}$ & 73.0 & 83.8 & 81.6  \\
     \bottomrule
    \end{tabular}
    }
\end{table}

\textbf{Study on the Dynamic Language Update Module.} 
To investigate the impact of language update frequency on DUTrack, we design the following experiment: building on the integration of the DTCM, we introduce language information and dynamically update it. In this set of experiments, we selected the variables \(\Delta S\) and \(\Delta D\) from Eq.~\ref{eq:s} and Eq.~\ref{eq:d} as the factors to study. 
The update strategy is as follows: if \(\Delta S\) is smaller than the set threshold, or \(\Delta D\) is larger than the set threshold, the language description will be updated. Here, the stride refers to the side length of the \(16\times16\) patches into which the image is divided.
In Tab.~\ref{tab:study_DL}, \#1 represents using only the initial language annotations. Comparing this with \#4 in Tab.~\ref{tab:study_DC}, we can see that the introduction of language information into the tracker results in improvements of 0.7\%, 1.3\%, and 2.3\% in AUC, P$_{\rm{Norm}}$, and P, respectively.
By comparing \#2, \#3, and \#4 in Tab.~\ref{tab:study_DL}, we can observe that as the update frequency increases, the AUC improves from 72.5\% to 73.0\%. This trend indicates that updating the language descriptions is highly effective in reducing the discrepancy between the language and visual modalities, significantly enhancing the performance of the vision-language tracker.
\begin{figure}
    \centering
    \includegraphics[width=0.48\textwidth]{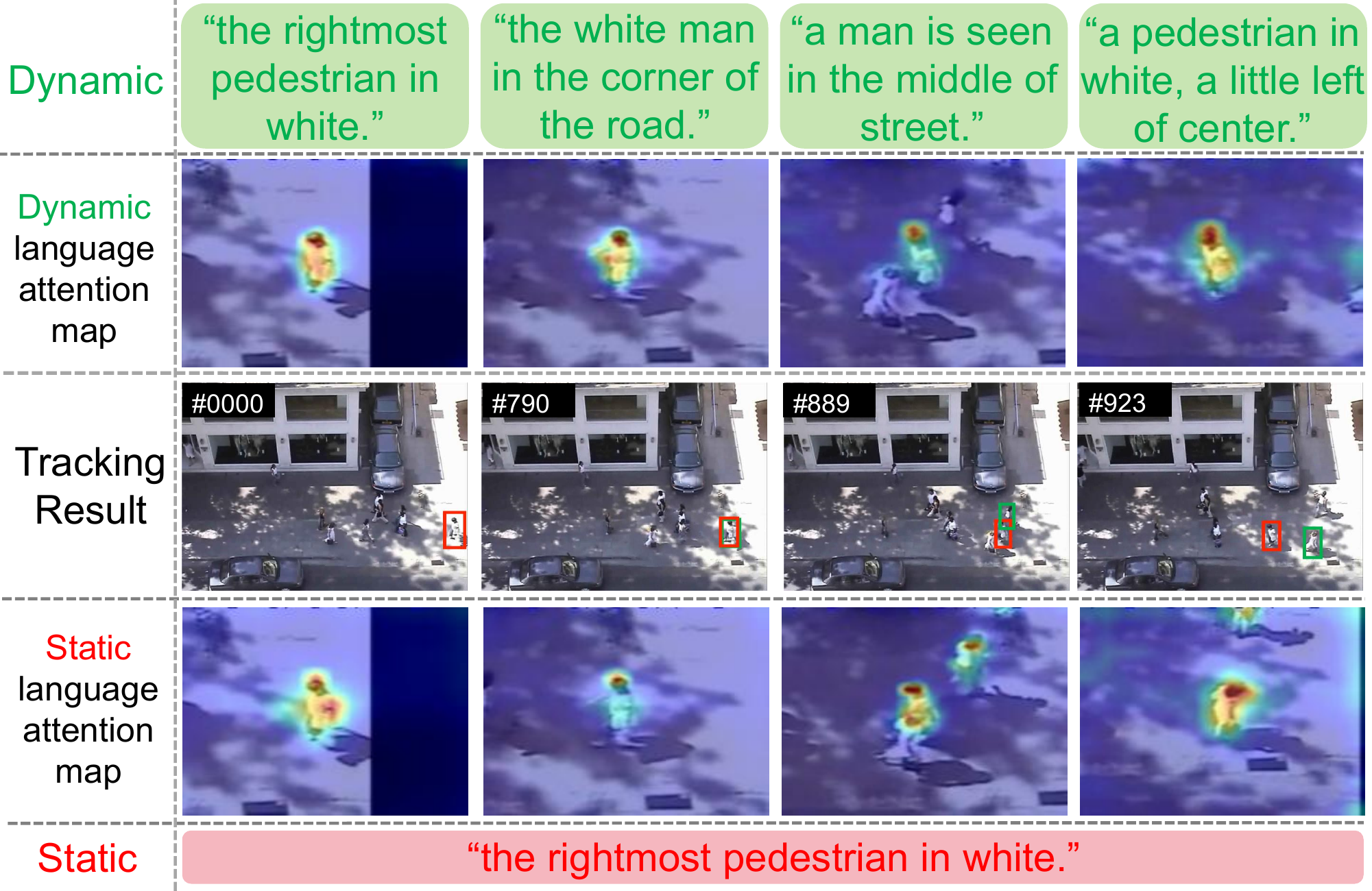}
    \caption{Visualization of the attention map of the Cls token on the search region using dynamic and initial language annotations. Red represents dynamic annotations, green represents static annotations.} 
    \label{fig:vis_attn}
\end{figure}

\begin{table}[h]
   
    \centering
    \caption{Study on the different LLM for DULM.}
    \label{tab:study_LLM}
    \fontsize{8}{6}\selectfont
    \resizebox{0.48\textwidth}{!}
    {
    \begin{tabular}{c|c|ccc}
     \toprule
     \multicolumn{1}{c|}{\multirow{2}{*}{Num}}&\multicolumn{1}{c|}{\multirow{2}{*}{Type}}& \multicolumn{3}{c}{LaSOT} 
       \\ \cline{3-5}
     &&AUC&P$_{\rm{Norm}}$ &P \\
     \midrule
     \#1&BLIP  & 73.0 & 83.8 & 81.6  \\
     \#2&BLIP-2  & 73.2  & 84.1 & 81.7  \\
     \#3&DTLLM-Concise  & 72.9 & 83.9 & 81.2 \\
     \#4&DTLLM-Detailed  & 72.5 & 83.3 & 80.6  \\
     \bottomrule
    \end{tabular}
    }
    \label{tab:LLM}
\end{table}

\textbf{Study on the LLM.} 
To investigate the impact of LLM, we conduct detailed experiments based on the current best results. The variable in this set of experiments is the type of LLM, and we test four different language generation methods.
As shown in Tab.~\ref{tab:study_LLM}, using BLIP \cite{blip} for language generation resulted in scores of AUC 73.0\%, P$_{\rm{Norm}}$ 83.8\%, and P 81.6\%. BLIP-2 \cite{blip-2}, used in \#2, improved AUC by 0.2\% compared to \#1. It can seamlessly map to the language model, significantly reducing computational costs while maintaining high performance. \#3 and \#4 use annotations generated by DTLLM \cite{dtllm} as language input, with the difference being the style of the generated language. The concise style annotations show results similar to \#1, but the detailed style annotations led to a drop in performance, with AUC and P$_{\rm{Norm}}$ decreasing by 0.5\% and P by 1.0\%. Our analysis suggests that the detailed annotations introduce too much learning pressure on the model, adding unnecessary noise that resulted in the performance decline.

\subsection{Visualization.}
To provide a more detailed demonstration of the effectiveness of our proposed method, Fig.~\ref{fig:eao_lasot} presents the AUC scores of DUTrack compared to current mainstream vision-language trackers and vision-only trackers across various challenges. Additionally, Fig.~\ref{fig:vis_box} visualizes the tracking results of three challenging sequences. Furthermore, in our experiments investigating DLUM, to verify the effectiveness of our proposed dynamic language annotation, we visualized the attention map of the \([CLS]\) token in the language description over the entire search area. As shown in Fig.~\ref{fig:vis_attn}, in the static language scenario, there is a noticeable misalignment between the tracker's attention and the tracking target. Although our dynamic language annotations may not provide highly precise semantic information, they eliminate this misalignment, thereby better assisting the tracker in locating the object.

\section{Conclusion}
In this paper, we introduce a new vision-language tracking framework, named DUTrack. Compared to previous vision-language trackers that rely solely on the initial template frame and language annotation, DUTrack addresses the issue of static multi-modal references deviating from the target's state by dynamically updating the multi-modal references. Extensive experiments demonstrate that, compared to previous SOTA vision-language trackers, our method effectively enhances vision-language tracking performance.

\textit{Limitation.} Our proposed multi-modal reference update method allows control over the update frequency by setting a threshold. Although manually setting the threshold is simple and reliable, it often requires extensive experimentation to find the optimal value, making the process time-consuming and challenging to achieve consistent results. Enhancing the flexibility and robustness of the update strategy is also our next research direction.

\section{Acknowledgments}
This work is supported by the Project of Guangxi Science and Technology (No.2024GXNSFGA010001), the National Natural Science Foundation of China (No.U23A20383, 62472109 and 62466051), the Guangxi "Young Bagui Scholar" Teams for Innovation and Research Project, the Research Project of Guangxi Normal University (No.2024DF001), the Innovation Project of Guangxi Graduate Education (YCBZ2024083), and the grant from Guangxi Colleges and Universities Key Laboratory of Intelligent Software (No.2024B01).
{
    \small
    \bibliographystyle{ieeenat_fullname}
    \bibliography{main}

\begin{thebibliography}{47}
\providecommand{\natexlab}[1]{#1}
\providecommand{\url}[1]{\texttt{#1}}
\expandafter\ifx\csname urlstyle\endcsname\relax
  \providecommand{\doi}[1]{doi: #1}\else
  \providecommand{\doi}{doi: \begingroup \urlstyle{rm}\Url}\fi

\bibitem[Alexey(2020)]{vit}
Dosovitskiy Alexey.
\newblock An image is worth 16x16 words: Transformers for image recognition at scale.
\newblock \emph{arXiv preprint arXiv: 2010.11929}, 2020.

\bibitem[Benchmark(2016)]{uav123}
UT Benchmark.
\newblock A benchmark and simulator for uav tracking.
\newblock In \emph{European Conference on Computer Vision}, 2016.

\bibitem[Bertinetto et~al.(2016)Bertinetto, Valmadre, Henriques, Vedaldi, and Torr]{simafc}
Luca Bertinetto, Jack Valmadre, Joao~F Henriques, Andrea Vedaldi, and Philip~HS Torr.
\newblock Fully-convolutional siamese networks for object tracking.
\newblock In \emph{Computer Vision--ECCV 2016 Workshops: Amsterdam, The Netherlands, October 8-10 and 15-16, 2016, Proceedings, Part II 14}, pages 850--865. Springer, 2016.

\bibitem[Chen et~al.(2021)Chen, Yan, Zhu, Wang, Yang, and Lu]{chen2021transformer}
Xin Chen, Bin Yan, Jiawen Zhu, Dong Wang, Xiaoyun Yang, and Huchuan Lu.
\newblock Transformer tracking.
\newblock In \emph{Proceedings of the IEEE/CVF conference on computer vision and pattern recognition}, pages 8126--8135, 2021.

\bibitem[Chen et~al.(2023)Chen, Peng, Wang, Lu, and Hu]{seqtrack}
Xin Chen, Houwen Peng, Dong Wang, Huchuan Lu, and Han Hu.
\newblock Seqtrack: Sequence to sequence learning for visual object tracking.
\newblock In \emph{Proceedings of the IEEE/CVF conference on computer vision and pattern recognition}, pages 14572--14581, 2023.

\bibitem[Chen et~al.(2020)Chen, Zhong, Li, Zhang, and Ji]{siamBAN}
Zedu Chen, Bineng Zhong, Guorong Li, Shengping Zhang, and Rongrong Ji.
\newblock Siamese box adaptive network for visual tracking.
\newblock In \emph{Proceedings of the IEEE/CVF conference on computer vision and pattern recognition}, pages 6668--6677, 2020.

\bibitem[Devlin(2018)]{bert}
Jacob Devlin.
\newblock Bert: Pre-training of deep bidirectional transformers for language understanding.
\newblock \emph{arXiv preprint arXiv:1810.04805}, 2018.

\bibitem[Fan et~al.(2019)Fan, Lin, Yang, Chu, Deng, Yu, Bai, Xu, Liao, and Ling]{LaSOT}
Heng Fan, Liting Lin, Fan Yang, Peng Chu, Ge Deng, Sijia Yu, Hexin Bai, Yong Xu, Chunyuan Liao, and Haibin Ling.
\newblock Lasot: A high-quality benchmark for large-scale single object tracking.
\newblock In \emph{Proceedings of the IEEE/CVF conference on computer vision and pattern recognition}, pages 5374--5383, 2019.

\bibitem[Fan et~al.(2021)Fan, Bai, Lin, Yang, Chu, Deng, Yu, Harshit, Huang, Liu, et~al.]{LaSOText}
Heng Fan, Hexin Bai, Liting Lin, Fan Yang, Peng Chu, Ge Deng, Sijia Yu, Harshit, Mingzhen Huang, Juehuan Liu, et~al.
\newblock Lasot: A high-quality large-scale single object tracking benchmark.
\newblock \emph{International Journal of Computer Vision}, 129:\penalty0 439--461, 2021.

\bibitem[Feng et~al.(2020{\natexlab{a}})Feng, Ablavsky, Bai, Li, and Sclaroff]{feng2020}
Qi Feng, Vitaly Ablavsky, Qinxun Bai, Guorong Li, and Stan Sclaroff.
\newblock Real-time visual object tracking with natural language description.
\newblock In \emph{Proceedings of the IEEE/CVF winter conference on applications of computer vision}, pages 700--709, 2020{\natexlab{a}}.

\bibitem[Feng et~al.(2020{\natexlab{b}})Feng, Ablavsky, Bai, Li, and Sclaroff]{lstmtrack}
Qi Feng, Vitaly Ablavsky, Qinxun Bai, Guorong Li, and Stan Sclaroff.
\newblock Real-time visual object tracking with natural language description.
\newblock In \emph{{WACV}}, pages 689--698, 2020{\natexlab{b}}.

\bibitem[Feng et~al.(2021{\natexlab{a}})Feng, Ablavsky, Bai, and Sclaroff]{feng2021siamese}
Qi Feng, Vitaly Ablavsky, Qinxun Bai, and Stan Sclaroff.
\newblock Siamese natural language tracker: Tracking by natural language descriptions with siamese trackers.
\newblock In \emph{Proceedings of the IEEE/CVF conference on computer vision and pattern recognition}, pages 5851--5860, 2021{\natexlab{a}}.

\bibitem[Feng et~al.(2021{\natexlab{b}})Feng, Ablavsky, Bai, and Sclaroff]{siamesevl}
Qi Feng, Vitaly Ablavsky, Qinxun Bai, and Stan Sclaroff.
\newblock Siamese natural language tracker: Tracking by natural language descriptions with siamese trackers.
\newblock In \emph{Proceedings of the IEEE/CVF Conference on Computer Vision and Pattern Recognition}, pages 5851--5860, 2021{\natexlab{b}}.

\bibitem[Ge et~al.()Ge, Cao, Zhu, Zhang, Liu, Wang, and Liu]{consistencies}
Jiawei Ge, Jiuxin Cao, Xuelin Zhu, Xinyu Zhang, Chang Liu, Kun Wang, and Bo Liu.
\newblock Consistencies are all you need for semi-supervised vision-language tracking.
\newblock In \emph{ACM Multimedia 2024}.

\bibitem[Ge et~al.(2024)Ge, Cao, Zhu, Zhang, Liu, Wang, and Liu]{ATTrak}
Jiawei Ge, Jiuxin Cao, Xuelin Zhu, Xinyu Zhang, Chang Liu, Kun Wang, and Bo Liu.
\newblock Consistencies are all you need for semi-supervised vision-language tracking.
\newblock In \emph{ACM Multimedia 2024}, 2024.

\bibitem[Guo et~al.(2022{\natexlab{a}})Guo, Zhang, Fan, and Jing]{divertmore}
Mingzhe Guo, Zhipeng Zhang, Heng Fan, and Liping Jing.
\newblock Divert more attention to vision-language tracking.
\newblock \emph{Advances in Neural Information Processing Systems}, 35:\penalty0 4446--4460, 2022{\natexlab{a}}.

\bibitem[Guo et~al.(2022{\natexlab{b}})Guo, Zhang, Fan, Jing, Lyu, Li, and Hu]{TransInMo}
Mingzhe Guo, Zhipeng Zhang, Heng Fan, Liping Jing, Yilin Lyu, Bing Li, and Weiming Hu.
\newblock Learning target-aware representation for visual tracking via informative interactions.
\newblock In \emph{{IJCAI}}, pages 927--934, 2022{\natexlab{b}}.

\bibitem[Hu et~al.(2023)Hu, Zhong, Liang, Zhang, Li, Li, and Ji]{HTtrack}
Xiantao Hu, Bineng Zhong, Qihua Liang, Shengping Zhang, Ning Li, Xianxian Li, and Rongrong Ji.
\newblock Transformer tracking via frequency fusion.
\newblock \emph{IEEE Transactions on Circuits and Systems for Video Technology}, 34\penalty0 (2):\penalty0 1020--1031, 2023.

\bibitem[Hu et~al.(2024{\natexlab{a}})Hu, Tai, Zhao, Zhao, Zhang, Li, Zhong, and Yang]{hu2}
Xiantao Hu, Ying Tai, Xu Zhao, Chen Zhao, Zhenyu Zhang, Jun Li, Bineng Zhong, and Jian Yang.
\newblock Exploiting multimodal spatial-temporal patterns for video object tracking.
\newblock \emph{arXiv preprint arXiv:2412.15691}, 2024{\natexlab{a}}.

\bibitem[Hu et~al.(2024{\natexlab{b}})Hu, Zhong, Liang, Zhang, Li, and Li]{hu1}
Xiantao Hu, Bineng Zhong, Qihua Liang, Shengping Zhang, Ning Li, and Xianxian Li.
\newblock Toward modalities correlation for rgb-t tracking.
\newblock \emph{IEEE Transactions on Circuits and Systems for Video Technology}, 34\penalty0 (10):\penalty0 9102--9111, 2024{\natexlab{b}}.

\bibitem[Huang et~al.(2019)Huang, Zhao, and Huang]{got10k}
Lianghua Huang, Xin Zhao, and Kaiqi Huang.
\newblock Got-10k: A large high-diversity benchmark for generic object tracking in the wild.
\newblock \emph{IEEE transactions on pattern analysis and machine intelligence}, 43\penalty0 (5):\penalty0 1562--1577, 2019.

\bibitem[Li et~al.(2022)Li, Li, Xiong, and Hoi]{blip}
Junnan Li, Dongxu Li, Caiming Xiong, and Steven Hoi.
\newblock Blip: Bootstrapping language-image pre-training for unified vision-language understanding and generation.
\newblock In \emph{International conference on machine learning}, pages 12888--12900. PMLR, 2022.

\bibitem[Li et~al.(2023)Li, Li, Savarese, and Hoi]{blip-2}
Junnan Li, Dongxu Li, Silvio Savarese, and Steven Hoi.
\newblock Blip-2: Bootstrapping language-image pre-training with frozen image encoders and large language models.
\newblock In \emph{International conference on machine learning}, pages 19730--19742. PMLR, 2023.

\bibitem[Li et~al.(2024)Li, Feng, Hu, Wu, Zhang, Zhang, and Huang]{dtllm}
Xuchen Li, Xiaokun Feng, Shiyu Hu, Meiqi Wu, Dailing Zhang, Jing Zhang, and Kaiqi Huang.
\newblock Dtllm-vlt: Diverse text generation for visual language tracking based on llm.
\newblock In \emph{Proceedings of the IEEE/CVF Conference on Computer Vision and Pattern Recognition}, pages 7283--7292, 2024.

\bibitem[Li et~al.(2017)Li, Tao, Gavves, Snoek, and Smeulders]{otblang}
Zhenyang Li, Ran Tao, Efstratios Gavves, Cees~GM Snoek, and Arnold~WM Smeulders.
\newblock Tracking by natural language specification.
\newblock In \emph{Proceedings of the IEEE conference on computer vision and pattern recognition}, pages 6495--6503, 2017.

\bibitem[Lin et~al.(2014)Lin, Maire, Belongie, Hays, Perona, Ramanan, Doll{\'a}r, and Zitnick]{coco}
Tsung-Yi Lin, Michael Maire, Serge Belongie, James Hays, Pietro Perona, Deva Ramanan, Piotr Doll{\'a}r, and C~Lawrence Zitnick.
\newblock Microsoft coco: Common objects in context.
\newblock In \emph{Computer Vision--ECCV 2014: 13th European Conference, Zurich, Switzerland, September 6-12, 2014, Proceedings, Part V 13}, pages 740--755. Springer, 2014.

\bibitem[Ma et~al.(2024)Ma, Tang, Yang, Zhang, Zhang, and Kang]{UVLTrack}
Yinchao Ma, Yuyang Tang, Wenfei Yang, Tianzhu Zhang, Jinpeng Zhang, and Mengxue Kang.
\newblock Unifying visual and vision-language tracking via contrastive learning.
\newblock In \emph{Proceedings of the AAAI Conference on Artificial Intelligence}, pages 4107--4116, 2024.

\bibitem[Muller et~al.(2018)Muller, Bibi, Giancola, Alsubaihi, and Ghanem]{trackingnet}
Matthias Muller, Adel Bibi, Silvio Giancola, Salman Alsubaihi, and Bernard Ghanem.
\newblock Trackingnet: A large-scale dataset and benchmark for object tracking in the wild.
\newblock In \emph{Proceedings of the European conference on computer vision (ECCV)}, pages 300--317, 2018.

\bibitem[Shao et~al.(2024)Shao, He, Ye, Feng, Luo, and Chen]{queryNLT}
Yanyan Shao, Shuting He, Qi Ye, Yuchao Feng, Wenhan Luo, and Jiming Chen.
\newblock Context-aware integration of language and visual references for natural language tracking.
\newblock In \emph{Proceedings of the IEEE/CVF Conference on Computer Vision and Pattern Recognition}, pages 19208--19217, 2024.

\bibitem[Shi et~al.(2024)Shi, Zhong, Liang, Li, Zhang, and Li]{evpt}
Liangtao Shi, Bineng Zhong, Qihua Liang, Ning Li, Shengping Zhang, and Xianxian Li.
\newblock Explicit visual prompts for visual object tracking.
\newblock In \emph{Proceedings of the AAAI Conference on Artificial Intelligence}, pages 4838--4846, 2024.

\bibitem[Tian et~al.(2024)Tian, Xie, Qiu, Jiao, Wang, Tian, and Ye]{fast-itpn}
Yunjie Tian, Lingxi Xie, Jihao Qiu, Jianbin Jiao, Yaowei Wang, Qi Tian, and Qixiang Ye.
\newblock Fast-itpn: Integrally pre-trained transformer pyramid network with token migration.
\newblock \emph{IEEE Transactions on Pattern Analysis and Machine Intelligence}, 2024.

\bibitem[Wang et~al.(2021)Wang, Shu, Zhang, Jiang, Wang, Tian, and Wu]{tnl2k}
Xiao Wang, Xiujun Shu, Zhipeng Zhang, Bo Jiang, Yaowei Wang, Yonghong Tian, and Feng Wu.
\newblock Towards more flexible and accurate object tracking with natural language: Algorithms and benchmark.
\newblock In \emph{Proceedings of the IEEE/CVF conference on computer vision and pattern recognition}, pages 13763--13773, 2021.

\bibitem[Xie et~al.(2024)Xie, Zhong, Mo, Zhang, Shi, Song, and Ji]{aqatrack}
Jinxia Xie, Bineng Zhong, Zhiyi Mo, Shengping Zhang, Liangtao Shi, Shuxiang Song, and Rongrong Ji.
\newblock Autoregressive queries for adaptive tracking with spatio-temporal transformers.
\newblock In \emph{Proceedings of the IEEE/CVF Conference on Computer Vision and Pattern Recognition}, pages 19300--19309, 2024.

\bibitem[Yan et~al.(2021)Yan, Peng, Fu, Wang, and Lu]{stark}
Bin Yan, Houwen Peng, Jianlong Fu, Dong Wang, and Huchuan Lu.
\newblock Learning spatio-temporal transformer for visual tracking.
\newblock In \emph{Proceedings of the IEEE/CVF international conference on computer vision}, pages 10448--10457, 2021.

\bibitem[Yang et~al.(2021)Yang, Kumar, Chen, Su, and Luo]{GTI}
Zhengyuan Yang, Tushar Kumar, Tianlang Chen, Jingsong Su, and Jiebo Luo.
\newblock Grounding-tracking-integration.
\newblock \emph{{IEEE} Trans. Circuits Syst. Video Technol.}, pages 3433--3443, 2021.

\bibitem[Ye et~al.(2022)Ye, Chang, Ma, Shan, and Chen]{ostrack}
Botao Ye, Hong Chang, Bingpeng Ma, Shiguang Shan, and Xilin Chen.
\newblock Joint feature learning and relation modeling for tracking: A one-stream framework.
\newblock In \emph{European Conference on Computer Vision}, pages 341--357. Springer, 2022.

\bibitem[Zhang et~al.(2023{\natexlab{a}})Zhang, Sun, Yang, Liu, Liu, Zhou, and Wang]{allinone}
Chunhui Zhang, Xin Sun, Yiqian Yang, Li Liu, Qiong Liu, Xi Zhou, and Yanfeng Wang.
\newblock All in one: Exploring unified vision-language tracking with multi-modal alignment.
\newblock In \emph{Proceedings of the 31st ACM International Conference on Multimedia}, pages 5552--5561, 2023{\natexlab{a}}.

\bibitem[Zhang et~al.(2024)Zhang, Zhong, Liang, Mo, Li, and Song]{one_zhang}
Guangtong Zhang, Bineng Zhong, Qihua Liang, Zhiyi Mo, Ning Li, and Shuxiang Song.
\newblock One-stream stepwise decreasing for vision-language tracking.
\newblock \emph{IEEE Transactions on Circuits and Systems for Video Technology}, pages 1--1, 2024.

\bibitem[Zhang et~al.(2023{\natexlab{b}})Zhang, Wang, Zhang, Zhang, and Zhong]{onevl}
Huanlong Zhang, Jingchao Wang, Jianwei Zhang, Tianzhu Zhang, and Bineng Zhong.
\newblock One-stream vision-language memory network for object tracking.
\newblock \emph{IEEE Transactions on Multimedia}, 26:\penalty0 1720--1730, 2023{\natexlab{b}}.

\bibitem[Zhang et~al.(2019)Zhang, Gonzalez-Garcia, Weijer, Danelljan, and Khan]{updataNet}
Lichao Zhang, Abel Gonzalez-Garcia, Joost Van~De Weijer, Martin Danelljan, and Fahad~Shahbaz Khan.
\newblock Learning the model update for siamese trackers.
\newblock In \emph{Proceedings of the IEEE/CVF international conference on computer vision}, pages 4010--4019, 2019.

\bibitem[Zhang et~al.(2022)Zhang, Wang, and Liang]{siamrdt}
Qian Zhang, Zihao Wang, and Hong Liang.
\newblock Siamrdt: An object tracking algorithm based on a reliable dynamic template.
\newblock \emph{Symmetry}, 14\penalty0 (4):\penalty0 762, 2022.

\bibitem[Zhang et~al.(2023{\natexlab{c}})Zhang, Tian, Xie, Huang, Dai, Ye, and Tian]{hivit}
Xiaosong Zhang, Yunjie Tian, Lingxi Xie, Wei Huang, Qi Dai, Qixiang Ye, and Qi Tian.
\newblock Hivit: A simpler and more efficient design of hierarchical vision transformer.
\newblock In \emph{The Eleventh International Conference on Learning Representations}, 2023{\natexlab{c}}.

\bibitem[Zhao et~al.(2023)Zhao, Wang, Wang, Lu, and Ruan]{transvl}
Haojie Zhao, Xiao Wang, Dong Wang, Huchuan Lu, and Xiang Ruan.
\newblock Transformer vision-language tracking via proxy token guided cross-modal fusion.
\newblock \emph{Pattern Recognit. Lett.}, 168:\penalty0 10--16, 2023.

\bibitem[Zheng et~al.(2023)Zheng, Zhong, Liang, Li, Ji, and Li]{mmtrack}
Yaozong Zheng, Bineng Zhong, Qihua Liang, Guorong Li, Rongrong Ji, and Xianxian Li.
\newblock Towards unified token learning for vision-language tracking.
\newblock \emph{IEEE Transactions on Circuits and Systems for Video Technology}, 2023.

\bibitem[Zheng et~al.(2024)Zheng, Zhong, Liang, Mo, Zhang, and Li]{odtrack}
Yaozong Zheng, Bineng Zhong, Qihua Liang, Zhiyi Mo, Shengping Zhang, and Xianxian Li.
\newblock Odtrack: Online dense temporal token learning for visual tracking.
\newblock In \emph{Proceedings of the AAAI Conference on Artificial Intelligence}, pages 7588--7596, 2024.

\bibitem[Zhou et~al.(2023)Zhou, Zhou, Mao, and He]{JointVL}
Li Zhou, Zikun Zhou, Kaige Mao, and Zhenyu He.
\newblock Joint visual grounding and tracking with natural language specification.
\newblock In \emph{Proceedings of the IEEE/CVF Conference on Computer Vision and Pattern Recognition}, pages 23151--23160, 2023.

\bibitem[Zhou et~al.(2022)Zhou, Chen, Pei, Mao, Wang, and He]{GTELT}
Zikun Zhou, Jianqiu Chen, Wenjie Pei, Kaige Mao, Hongpeng Wang, and Zhenyu He.
\newblock Global tracking via ensemble of local trackers.
\newblock In \emph{{CVPR}}, pages 8751--8760, 2022.

\end{thebibliography}
}


\end{document}